\documentclass[]{spie}  

\usepackage{amsmath,amsfonts,amssymb}
\usepackage{graphicx}
\usepackage[colorlinks=true, allcolors=blue]{hyperref}

\title{Crowd-Assisted Polyp Annotation of Virtual Colonoscopy Videos}

\author[1]{Ji Hwan Park}
\author[1]{Saad Nadeem}
\author[1]{Joseph Marino}
\author[2]{Kevin Baker}
\author[2]{Matthew Barish}
\author[1]{Arie Kaufman}
\affil[1]{Department of Computer Science, Stony Brook University, Stony Brook, NY, USA}
\affil[2]{Department of Radiology, Stony Brook Medicine, Stony Brook NY, USA}

\authorinfo{Further author information: \\E-mail J.H.P., S.N., J.M., A.K.: \{ jihwpark $|$ sanadeem $|$ jmarino $|$ ari \}@cs.stonybrook.edu\\
E-mail K.B., M.B.: \{ Kevin.Baker $|$ Matthew.Barish \}@stonybrookmedicine.edu}

\begin{document} 
\maketitle

\begin{abstract}
Virtual colonoscopy (VC) allows a radiologist to navigate through a 3D colon model reconstructed from a computed tomography scan of the abdomen, looking for polyps, the precursors of colon cancer. Polyps are seen as protrusions on the colon wall and haustral folds, visible in the VC fly-through videos. A complete review of the colon surface requires full navigation from the rectum to the cecum in antegrade and retrograde directions, which is a tedious task that takes an average of 30 minutes. Crowdsourcing is a technique for non-expert users to perform certain tasks, such as image or video annotation. In this work, we use crowdsourcing for the examination of complete VC fly-through videos for polyp annotation by non-experts.  The motivation for this is to potentially help the radiologist reach a diagnosis in a shorter period of time, and provide a stronger confirmation of the eventual diagnosis. The crowdsourcing interface includes an interactive tool for the crowd to annotate suspected polyps in the video with an enclosing box. Using our workflow, we achieve an overall polyps-per-patient sensitivity of 87.88\% (95.65\% for polyps $\geq$5mm and 70\% for polyps $<$5mm). We also demonstrate the efficacy and effectiveness of a non-expert user in detecting and annotating polyps and discuss their possibility in aiding radiologists in VC examinations.  
\end{abstract}

\keywords{Virtual colonoscopy, crowdsourcing, colonic polyps}

\section{INTRODUCTION}
\label{sec:intro}  

Virtual colonoscopy (VC) employs CT scanning of the patient's abdomen together with advanced visualization techniques to virtually navigate within a reconstructed 3D colon model searching for colorectal polyps, the precursor of cancer. VC is widely recognized as a highly sensitive and specific test for identifying polyps in the colon~\cite{Johnson:2008}. It is a comfortable, inexpensive, very low risk, and fast procedure. One of the major issues with VC, though, is the required interpretation time by the radiologists. In order to ensure full coverage of the colon surface, a typical protocol is to acquire two scans of the patient in different positions (e.g., supine and prone) and to then virtually traverse each scan in both antegrade and retrograde directions. These multiple fly-throughs are used to ensure sufficient coverage of the colon surface, but are tedious and time consuming, as most regions of the colon are free of polyps. Computer-aided detection (CAD) algorithms have been developed to identify polyps automatically but these have false negative findings, missing lesions with adherent contrast medium, flat adenomas, and adenomas located on or adjacent to normal colonic folds. In order to increase sensitivity, specificity is usually sacrificed, thus increasing the number of false positive polyp candidates that must be rejected. This requires significant human intervention and time even though the majority of false positive candidates can be rejected by individuals with even minimal training.


Crowdsourcing seeks to engage the general masses, and previous studies have shown that most workers are forthright in their intentions~\cite{Suri:2011}. Crowdsourcing has been shown to be a good method for annotating endoscopic images. In the colon domain, there have been a few attempts to combine computer-aided detection with crowdsourcing to overcome some of the limitations~\cite{McKenna:2012,Wang:2011}. These initial attempts have shown that the crowd can be used to reject false positive polyp candidates, though little work has been done to determine if the crowd can be used for primary polyp detection or as a means of excluding the presence of a polyp within a segment of the colon. Other work has shown that workers who participated in two studies improved significantly in performance between the first and second study~\cite{nguyen:2012}. One small study was also performed where workers were asked to identify whether or not a short video segment contained a polyp, and the crowd was generally successful in identifying which segments contained a polyp~\cite{park:2017:spie}. 

In comparison to previous works, our study presents the crowd workers with complete fly-through videos and requires the workers to annotate the exact locations of polyps in the video frames.  Our goal is to see how well the crowd, as a whole, can be used for identifying specific polyps, and to see what amount of agreement amongst the crowd is needed to obtain good detection results.


\section{METHODS}
\label{sec:methods} 

\subsection{Video Generation}
We have used the commercially available FDA-approved Viatronix V3D-Colon virtual colonoscopy system to automatically generate four centerline fly-through videos for each patient VC dataset (from rectum to cecum and from cecum to rectum in both supine and prone orientations). We have only used VC datasets for which we can recover a complete single colon segment from both the supine and prone scans, since segmentation artifacts sometimes lead to multiple 3D colon segments from a single VC dataset. The videos are captured at 15 frames per second (fps) with a resolution of 256$\times$256 pixels and a 90$^{\circ}$ field-of-view.

\begin{figure*}[t]
    \centering
    \includegraphics[width=0.77\textwidth]{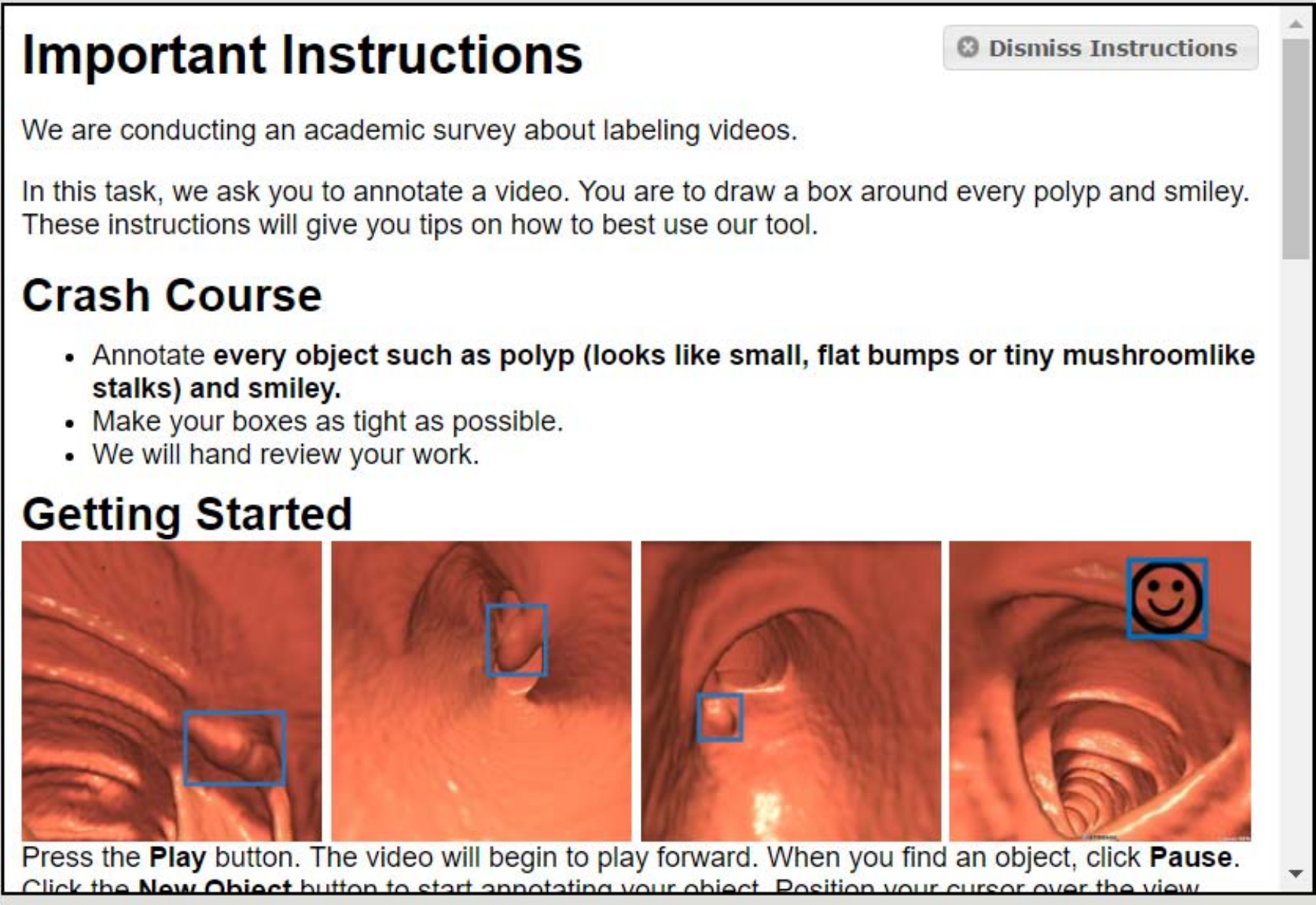}
    \caption{A portion of the instructions presented to each user, including examples of a variety of polyps and the quality control smiley.}
    \label{fig:instruction}
\end{figure*}

\subsection{Crowd Task}
To present our study to the non-expert crowd workers, we made use of the Amazon Mechanical Turk (MTurk) platform, which is a popular method of obtaining reliable crowd workers at modest cost while allowing us to reach a large and diverse population of users~\cite{paolacci:2010}. Our study was identified on the MTurk platform as an academic survey which would require video annotations. Our annotation tool is a modification of an existing open source crowdsourcing video annotation tool~\cite{Vondrick:2013}.

\begin{figure*}[tb]
    \centering \footnotesize
    \includegraphics[width=0.95\textwidth]{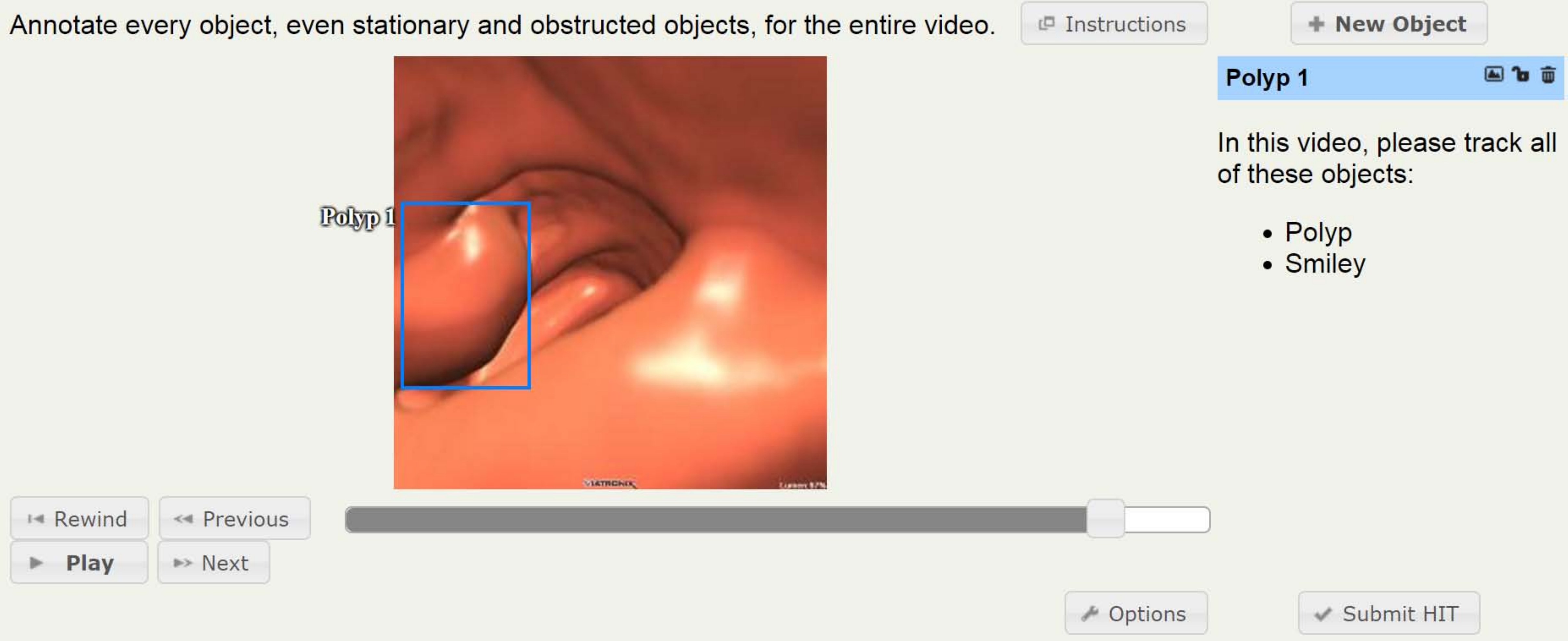} (a)
    \includegraphics[width=0.95\textwidth]{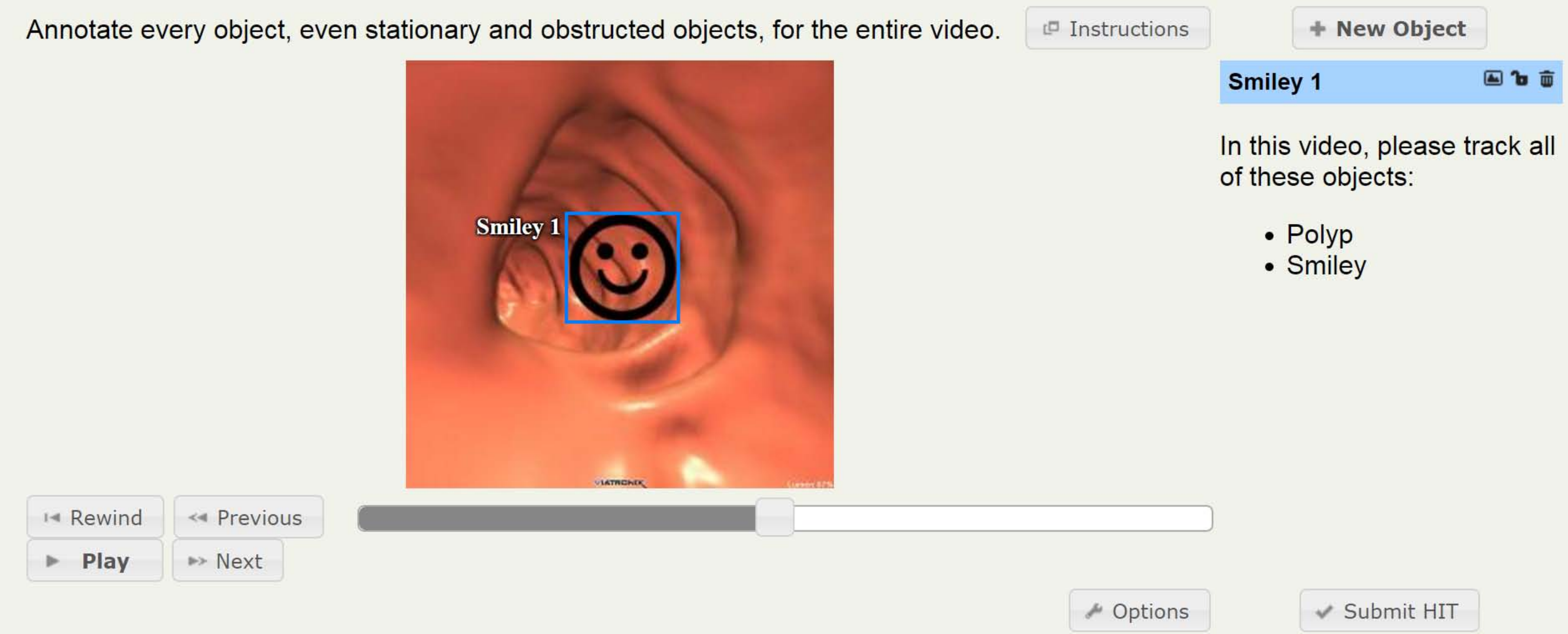} (b)
    \caption{Two views of the interface for our task, showing VC frames with (a) a polyp and (b) a smiley.}
    \label{fig:interface}
\end{figure*}

When users first select this human intelligence task (HIT), they are provided with brief instructions (see Figure~\ref{fig:instruction}) including the main objective of the study, how to use the system, some example polyps, and how the workers' results will be evaluated. While workers read the instructions, the video is downloaded in the background. Once the video download is complete, the workers can play, pause, and rewind the video at will. Additionally, they can step to the previous or next frame. When the workers find a polyp and/or smiley (for quality control purposes, see Section~\ref{sec:qc}), they first label it as a polyp or smiley and then annotate the polyp/smiley by drawing a rectangle around it.  Two examples of our interface with annotations are shown in Figure~\ref{fig:interface}, with one view displaying a polyp and the other showing a smiley.

\subsection{Quality Control}
\label{sec:qc}
Quality control is an essential component in crowdsourcing applications.
In our study, only workers who have an approval rate of greater than 95\% in the Amazon system were allowed to participate.
Additionally, we added control objects in each task to help detect spammers. The control objects should be easily answered by the crowd since their purpose is not to test the ability of the workers, but rather to detect spammers who do not attempt to perform the task properly. To this end, we added 5 smileys per video, each of which was created at a random location in a randomly selected frame. Each smiley appeared in the same location for 25 consecutive frames, and the workers are required to annotate the smileys in the same manner as polyps. Figure~\ref{fig:interface}(b) shows an example smiley in a video.

\subsection{Ground Truth}
We generated the ground truth to compare the crowd results with by marking polyps in the videos based on the expert radiologists' VC reports. We used the same annotation tool as the crowd workers to annotate polyps in the videos, where we annotated each polyp in every video frame it is visible in (as opposed to the workers, who annotated polyps in only a single frame).


\section{Experiments}
\label{sec:experiments} 

This study utilized VC data from 14 patients, yielding a total of 56 VC videos (antegrade and retrograde directions in both supine and prone scans) and thus a total of 56 unique HITs. These 14 patients contained a total of 33 polyps (3 flat, 14 sessile, and 16 pedunculated polyps), with the number of polyps per patient ranging from 1 to 5. Since not every polyp was visible in each of the four views, the 56 videos contained a total of 114 visible polyps. The datasets contained a selection of both large ($\geq5mm$) and small ($<5mm$) sized polyps. Of the 33 total polyps, 10 polyps were less than $5mm$ in diameter. The patients were chosen from an anonymized training database, and were ranked in difficulty from Level 1 (least difficult) to Level 4 (most difficult). Of the selected datasets, two were ranked as Level 1, three as Level 2, five as Level 3, and four as Level 4.

Each VC video was viewed by ten workers, except for two which were viewed by eleven workers. A worker was not allowed to complete the same task multiple times, but was allowed to complete multiple tasks, each with a different VC video. A total of 125 workers participated in our study. While most workers completed only a single task, several performed multiple tasks, including one worker who completed 45 tasks.

\section{RESULTS}
\label{sec:results} 

\begin{figure*}[t]
    \centering
    \includegraphics[width=0.32\textwidth]{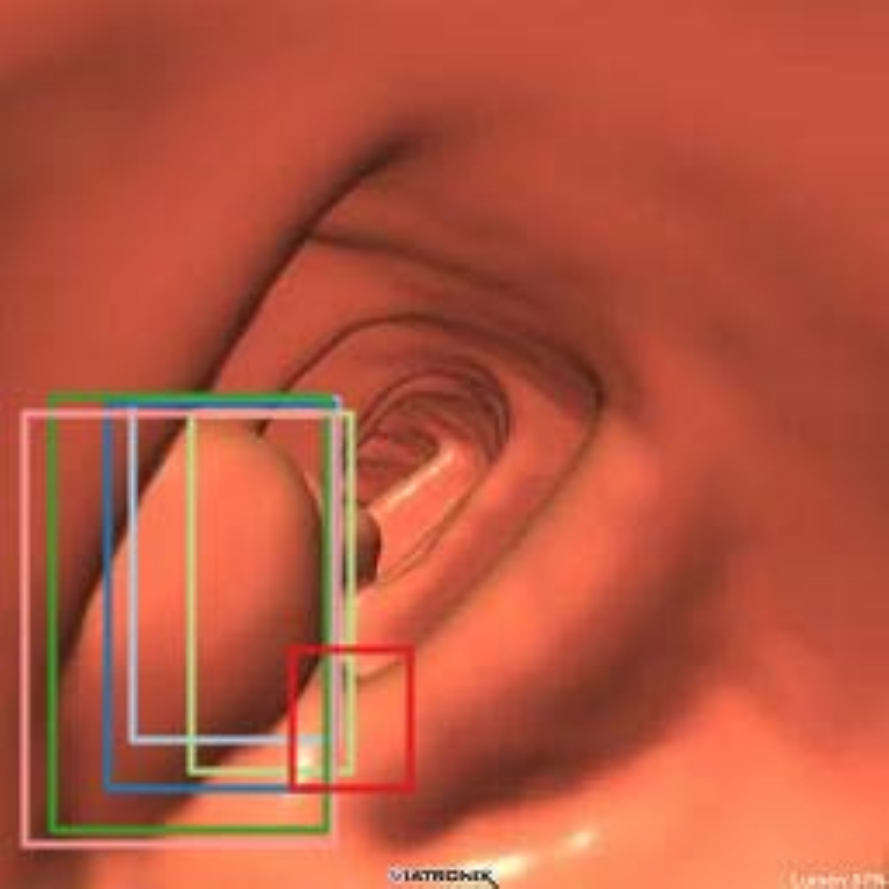}
    \includegraphics[width=0.32\textwidth]{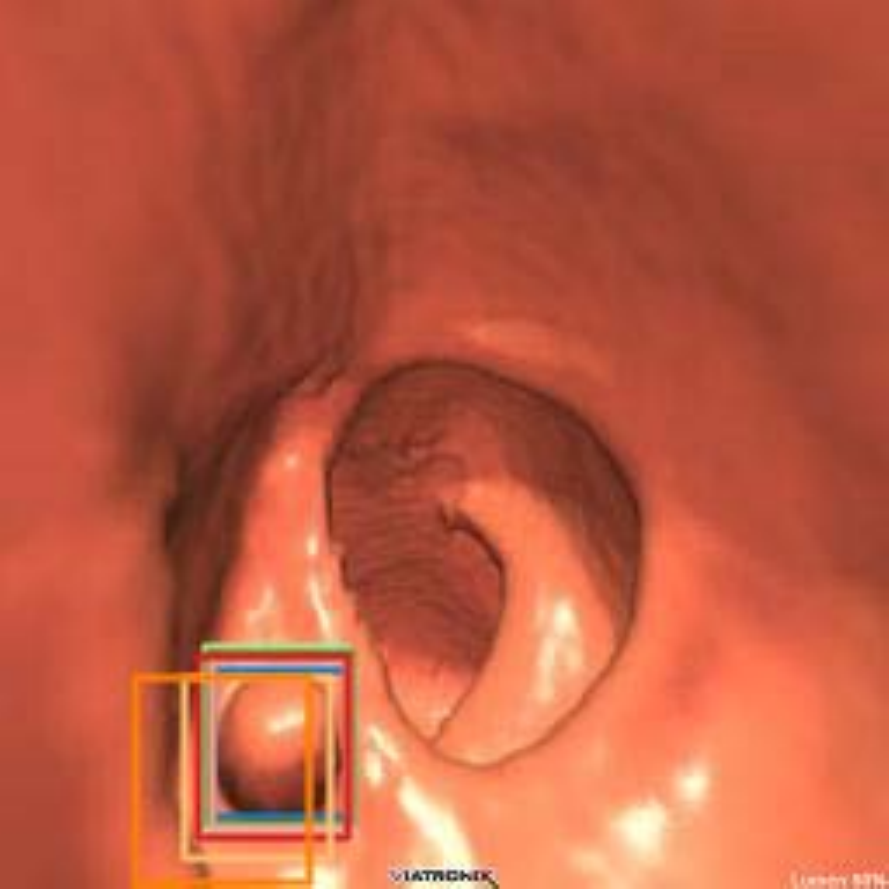}
    \includegraphics[width=0.32\textwidth]{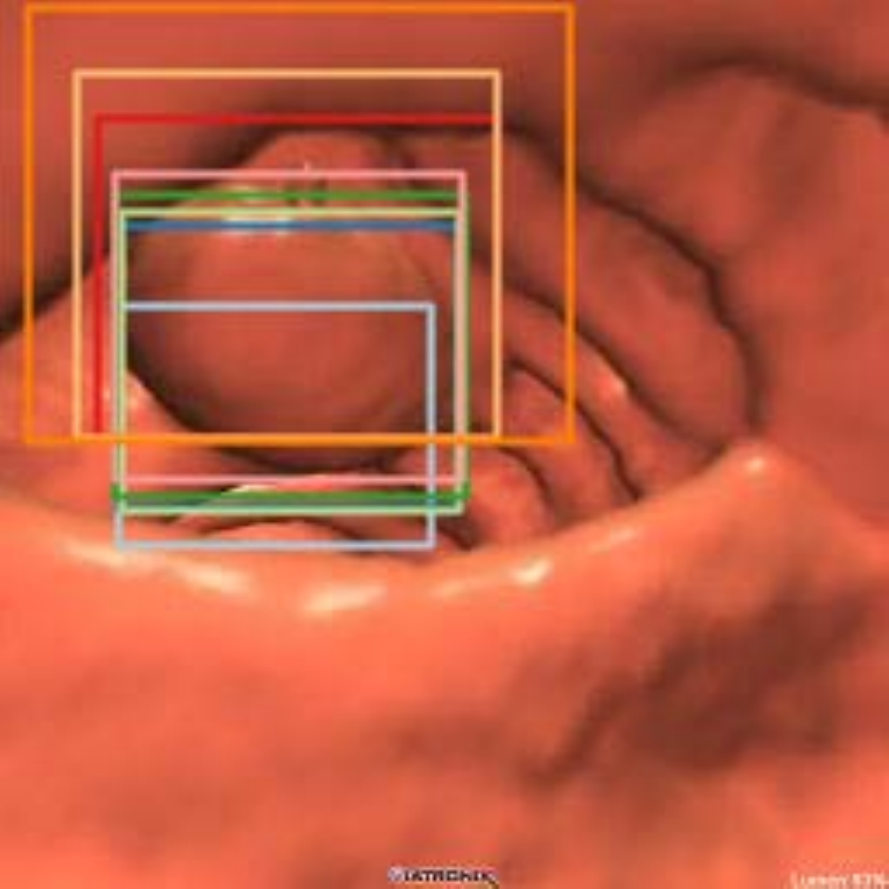}
    \caption{Examples of polyps in VC videos, and the corresponding annotations made by the crowd. All of the crowd's annotations for each polyp are superimposed here for illustrative purposes on a single selected frame containing the polyp, with each worker's annotation box in a different color.}
    \label{fig:annotations}
\end{figure*}

We have used the Dice similarity coefficient (DSC) to determine whether or not a worker annotation $W$ corresponds to an actual ground truth annotation $T$. The DSC is calculated as
\begin{equation}
    DSC_{WT} = \frac{ 2 \left| W \cap T \right| }{ \left| W \right| + \left| T \right| }
\end{equation}
and a worker annotation is considered a match to the ground truth annotation if $DSC_{WT}>0.5$. When considering if a polyp or a smiley is matched, only one frame of the polyp/smiley needs to be annotated by the crowd worker. Figure~\ref{fig:annotations} shows three examples of the various annotations made for three polyps from three different videos. Since the annotations were made in different frames of the original video, not all boxes are aligned tightly in the selected views shown here.

We have eliminated ``spammers" (or low quality workers) by eliminating tasks in which the worker did not match all five of the smileys that were inserted. This quality control measure removed 57 video tasks from the total number of 562 video tasks (54 HITs $\times$ 10 + 2 HITs $\times$ 11) that were performed in our study. Unless otherwise indicated, our discussion of the results considers only these remaining 505 tasks.

\begin{figure*}[t]
    \centering
    \includegraphics[width=0.48\textwidth]{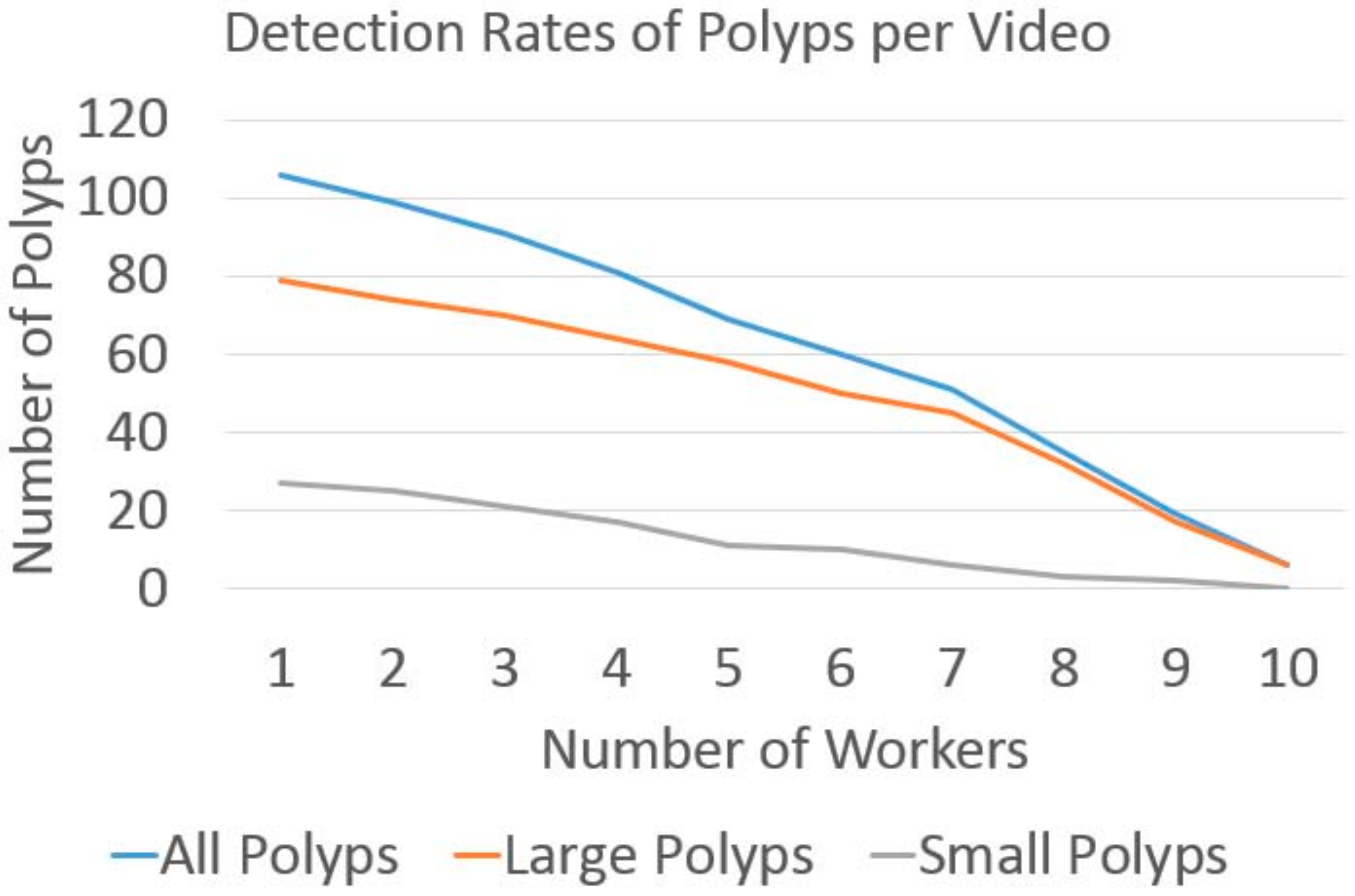}
    \hspace{2mm}
    \includegraphics[width=0.48\textwidth]{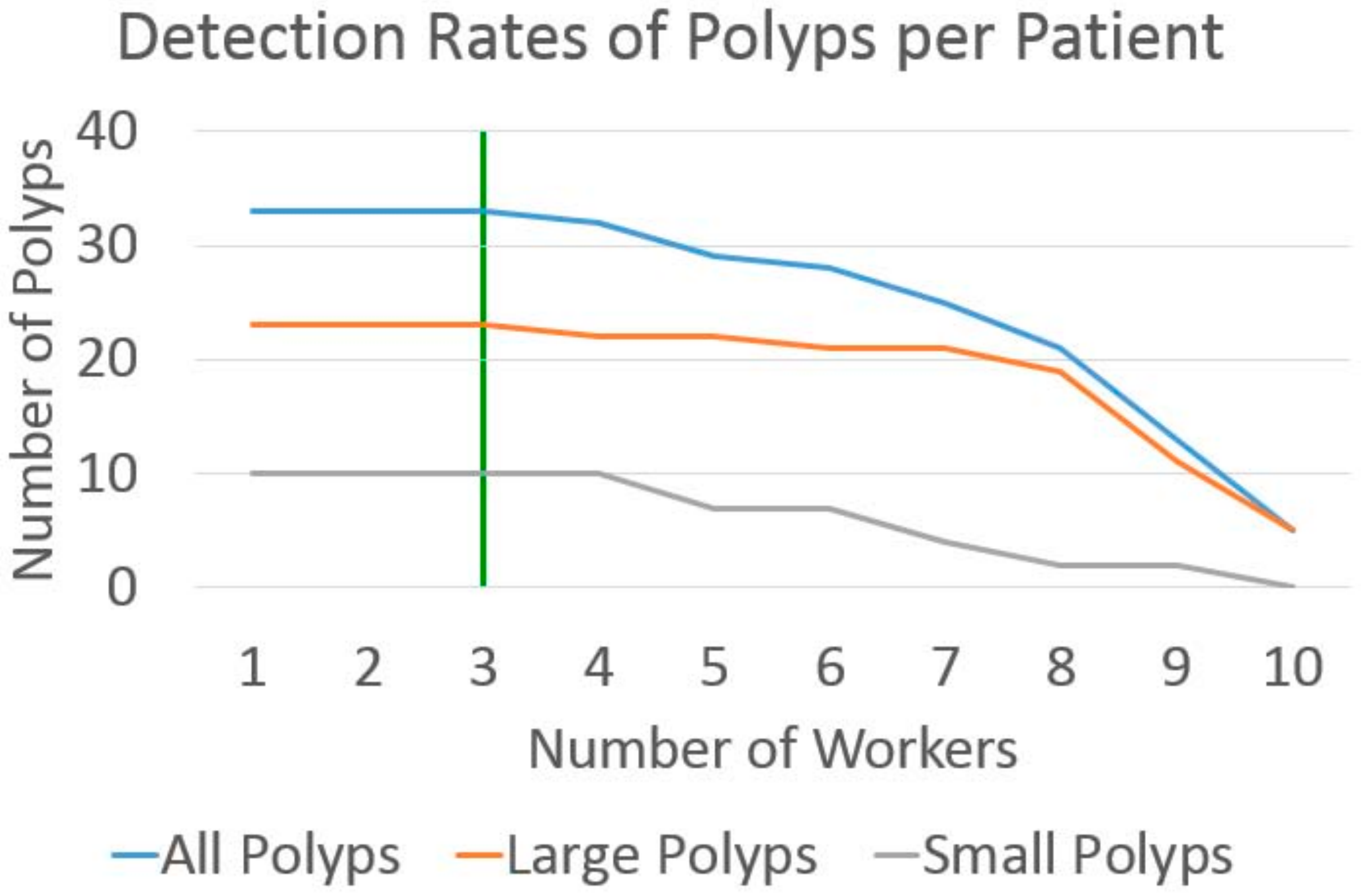}
    \caption{Detection rates based on the number of workers with a matching annotation required to consider a polyp found.  There were a total of 114 polyps (79 large and 35 small) on a per-video basis, and a total of 33 polyps (23 large and 10 small) on a per-patient basis. The green line in the per patient chart indicates the threshold for a 100\% polyp detection rate.}
    \label{fig:charts}
\end{figure*}

The accuracy of the crowd in properly annotating polyps is shown in Figure~\ref{fig:charts}. Broken down by individual video, our results showed that not every polyp was found by even a single worker; 106 of the 114 polyps were annotated by at least one worker. At the 50\% threshold, 69 of the polyps were annotated by at least five workers.

Since some polyps are not visible or are indistinct in a particular fly-through orientation, it is more important to look at the accuracy of the crowd in relation to physical polyps per patient (rather than per video, of which there are four videos generated per patient). Broken down by polyp per patient, our results show that all of the polyps were identified by at least three workers in a single video. By this, we mean that if a physical polyp was visible in multiple videos, at least three workers annotated the polyp in the same video; in most cases, the polyp was annotated accurately in multiple videos and by more than three workers. At the 50\% threshold, 29 of the 33 total polyps were annotated by at least five workers (22 of the 23 large polyps, and 7 of the 10 small polyps). This gives an overall polyps-per-patient sensitivity of 87.88\% (95.65\% for polyps $\geq$5mm and 70\% for polyps $<$5mm).

Removing the ``spammer'' workers reduced the total number of annotated frames resulting from the tasks. The 56 videos contained a total of 125,182 frames.  Using the results of all workers resulted in a total of 3,754 annotated frames. By removing the ``spammer'' workers, this number of annotated frames was reduced to 3,342, eliminating more than 10\% of the annotated frames and resulting in only 2.7\% of all frames in the original videos containing annotations.

\section{DISCUSSION}
\label{sec:discussion} 

\begin{figure*}[t]
    \centering
    \includegraphics[width=0.99\textwidth]{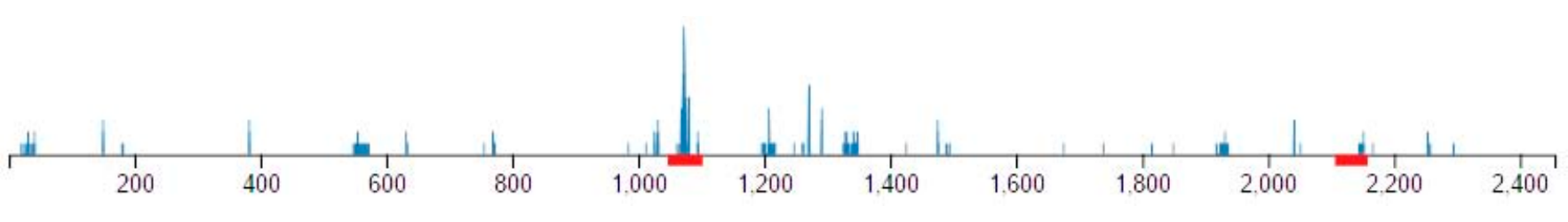}
    \includegraphics[width=0.99\textwidth]{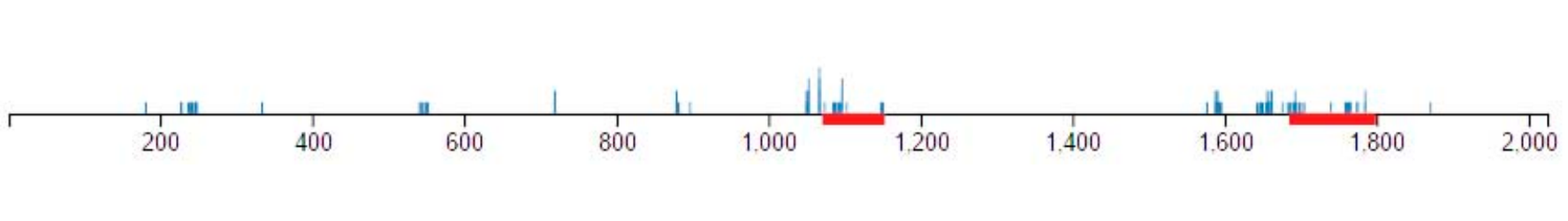}
    \includegraphics[width=0.99\textwidth]{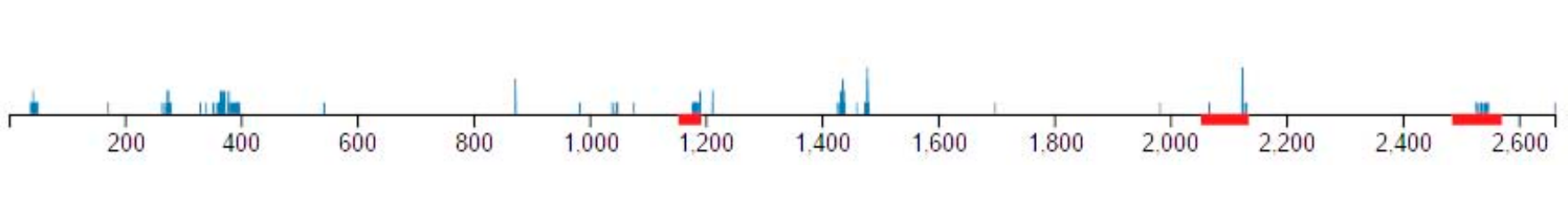}
    \caption{Timelines of three of the VC videos used in our study.  The blue bars above the line are worker annotation positions; the height of the line corresponds to the number of workers who annotated that specific frame.  The red regions below the line indicate the range of frames in which a ground truth polyp appears.}
    \label{fig:timelines}
\end{figure*}

We believe that the crowd results could be used in the future to assist radiologists in VC screenings.
Since we have only a few crowd-annotated frames per video, we can show these annotated frames to the expert radiologist, prior to the fly-through, who can quickly skim through these and discard the false positives due to issues such as segmentation artifacts or electronic cleansing problems. Moreover, the overlapping crowd annotations between successive frames can be grouped together to further reduce the total number of annotated frames. This can help control the automatic centerline flythrough by cleaning up the large swathes of regions with crowd-annotated false positives and allow the radiologist to speed through the regions void of polyps and slow down in regions with high-consensus crowd polyp annotations. Figure~\ref{fig:timelines} gives an example of where annotated frames are located within a colon video and in relation to the polyps, with non-annotated regions possibly being able to be reviewed more quickly by the radiologist in order to save time.

In addition to the insights gained regarding the ability of the crowd to locate polyps in VC videos, this study has also reinforced the importance of performing VC fly-throughs in both antegrade and retrograde directions in both the supine and prone scans.  In several instances, the crowd was successful in locating a polyp in one of these four views, but not in the other three.  Indeed, even for the ground truth annotations, three polyps were visible in only one view, two polyps were visible in only two views, and five polyps were visible in only three of the four views.

We believe that the crowd results could be further improved in future studies.  Most of the missed polyps fell into one of three categories: the polyp was occluded, the polyp was visible in only a few frames, or the polyp looked like a fold.  For occluded polyps, the multiple fly-throughs generally solve this issue.  In cases where the polyp is visible only briefly, adjusting the navigation speed to a slower rate might help to increase crowd performance in spotting these polyps.  Additional training and more examples of tricky polyps might also help improve the ability of the crowd in identifying polyps which look like folds.

\section{CONCLUSION}
\label{sec:conslusion} 
We have presented what is, to the best of our knowledge, the first crowdsourcing study in which non-expert crowd workers are requested to annotate colonic polyps directly from VC videos. Our results have shown that the crowd is quite adept at this task, even with only a modicum of training. We found that the crowd achieved an overall polyps-per-patient sensitivity of 87.88\% at the 50\% threshold (95.65\% for polyps $\geq$5mm and 70\% for polyps $<$5mm). Moreover, all polyps from each patient were identified in distinct videos by at least three workers. The total number of frames annotated by the workers was only 2.7\% of the total number of frames in the videos, yielding the potential to greatly reduce the required effort of the radiologist.

While this study was focused on determining whether or not crowd annotations of polyps was feasible, future work will focus on how the crowd results can be best used when the ground truth polyps are not known \emph{a priori}. Further research in this area will lead to a better understanding of how crowd results can be optimally utilized for assistance during VC examination.

\acknowledgments 
This work has been partially supported by the National Science Foundation grants IIP1069147, CNS1302246, IIS1527200, NRT1633299, and CNS1650499.


\end{document}